%% file: main.tex
\documentclass[runningheads]{llncs}
\usepackage{graphicx}
\usepackage{latexsym}
\usepackage{algorithm}
\usepackage{algorithmicx}
\usepackage[noend]{algpseudocode}
\usepackage{url}
\usepackage{amsmath}
\usepackage{amssymb}
\usepackage{amsfonts}
\usepackage{mathtools}
\usepackage{stmaryrd}
\usepackage{bm}
\usepackage{cite}
\usepackage{enumitem}
\usepackage{pifont}

\usepackage{xcolor}

\newcommand{\song}{\color{black}}

\DeclareMathOperator{\Enc}{\mathsf{Enc}}
\DeclareMathOperator{\Dec}{\mathsf{Dec}}

\DeclareMathOperator{\HomConv}{\mathsf{HomConv}}

\DeclareMathOperator{\Act}{\mathsf{SecAct}}
\DeclareMathOperator{\avgpool}{\mathsf{SecAvgPool}}
\DeclareMathOperator{\secargmax}{\mathsf{SecArgmax}}

\begin{document}

\title{
  BUNET: Blind Medical Image Segmentation Based on Secure UNET
}

\author{
Song Bian \inst{1} \and 
Xiaowei Xu \inst{2}  \and 
Weiwen Jiang \inst{3} \and
Yiyu Shi\inst{3} 
Takashi Sato\inst{1}
}
\authorrunning{S. Bian, et al.}
\institute{
Kyoto University\\
\email{\{sbian, takashi\}@easter.kuee.kyoto-u.ac.jp}
\and Guangdong Provincial People's Hospital\\
\email{xiao.wei.xu@foxmail.com}
\and University of Notre Dame\\
\email{{\{wjiang2, yshi4\}@nd.edu}}
}
\maketitle
\newcommand{\BoldVec}[1]{
  \expandafter\def\csname b#1\endcsname{{\bf{#1}}}%
}
\newcommand{\InnerProd}[2]{
  \expandafter\def\csname in#1#2\endcsname{\text{$\langle{\bf{#1}},{\bf{#2}}\rangle$}}%
}
\newcommand{\mRound}[1]{
  \text{$\left\lfloor #1\right\rceil$}%
}
\newcommand{\numberthis}{
  \addtocounter{equation}{1}\tag{\theequation}
}
\BoldVec{x}\BoldVec{y}\BoldVec{a}\BoldVec{s}\BoldVec{c}\BoldVec{r}\BoldVec{p}\BoldVec{e}\BoldVec{b}
\BoldVec{u}\BoldVec{v}\BoldVec{w}\BoldVec{q}\BoldVec{n}\BoldVec{t}\BoldVec{d}\BoldVec{z}\BoldVec{g}
\InnerProd{a}{s}
\newcommand{\ElementMult}[2]{
  #1\boxcircle #2
}

\def\zz{{\mathbb{Z}}}
\def\zr{{\mathbb{R}}}
\def\crr{{\mathcal{R}}}
\def\crk{{\mathcal{K}}}
\def\cra{{\mathcal{A}}}
\def\Conv{{\mathsf{Conv}}}
\def\FC{{\mathsf{FC}}}
\def\NTT{{\mathsf{NTT}}}
\def\INTT{{\mathsf{INTT}}}
\def\Perm{{\mathsf{Perm}}}
\def\rot{{\mathsf{rot}}}
\def\Decomp{{\mathsf{Decomp}}}
\def\bpt{b}
\def\etarot{\eta_{\rm rot}}
\def\etamult{\eta_{\rm mult}}
\def\aaprox{\alpha}
\def\fe{{f_{\rm E}}}
\def\fd{{f_{\rm D}}}
\def\bB{{\mathcal{B}}}
\def\ee{{\varepsilon}}
\def\bee{{\bm{\varepsilon}}}
\def\bxxi{{\bm{\xi}}}
\def\bpi{{\bm{\pi}}}
\def\etal{et~al.\ }
\def\clgq{\lceil\lg{q}\rceil}
\def\clgp{\lceil\lg{p}\rceil}
\algnewcommand\algorithmicforeach{\textbf{for each}}
\algdef{S}[FOR]{ForEach}[1]{\algorithmicforeach\ #1\ \algorithmicdo}
\def\bseg{\text{BUNET}}
\def\placeholder{NASS}
\def\SIMDScMult{{\mathsf{SIMDScMult}}}

\def\hatu{{\hat{u}}}
\def\hats{{\hat{s}}}
\def\hatw{{\hat{w}}}
\def\ovu{{\overline{u}}}
\def\ovw{{\overline{w}}}
\def\ovc{{\overline{c}}}
\def\ovbc{{\overline{\bc}}}
\def\ovbv{{\overline{\bv}}}
\def\ovbu{{\overline{\bu}}}
\def\ovbs{{\overline{\bs}}}
\def\hatU{{\hat{U}}}
\def\hatR{{\hat{R}}}
\def\hatW{{\hat{W}}}
\def\hatbu{{\hat{\bu}}}
\def\hatbr{{\hat{\br}}}
\def\hatbw{{\hat{\bw}}}
\def\hatbv{{\hat{\bv}}}
\def\hatbx{{\hat{\bx}}}
\def\hatby{{\hat{\by}}}
\def\hatbs{{\hat{\bs}}}
\def\bpt{b}
\def\etarot{\eta_{\rm rot}}
\def\etamult{\eta_{\rm mult}}
\def\aaprox{\alpha}
\def\fe{{f_{\rm E}}}
\def\fd{{f_{\rm D}}}
\def\ee{{\varepsilon}}
\def\bee{{\bm{\varepsilon}}}
\def\bpi{{\bm{\pi}}}
\def\etal{et~al.\ }
\def\clgq{\lceil\lg{q}\rceil}
\def\clgp{\lceil\lg{p}\rceil}
\def\pass{p_{\rm A}}
\def\share{\mathsf{Share}}
\def\rec{\mathsf{Rec}}
\def\networkparm{\mathsf{NetworkParms}}
\def\layerparm{\mathsf{LayerParms}}
\def\layerparml{\layerparm_{\rm L}}
\def\layerparmnl{\layerparm_{\rm NL}}

\sloppy
\begin{abstract}
\input{tex/abst.tex}
\end{abstract}

\setlength{\textfloatsep}{6pt}
\setlength{\floatsep}{6pt}
\setlength{\dbltextfloatsep}{6pt}

\input{tex/intro.tex}
\input{tex/background.tex}

\input{tex/propose.tex}

\input{tex/result.tex}
\input{tex/conclusion.tex}
\input{tex/ack.tex}

\bibliographystyle{splncs04}
\bibliography{cad,security,NAS,ai}

\input{appendix/appendix.tex}

\end{document}

%% file: tex/abst.tex
The strict security requirements placed on medical records by various privacy
regulations become major obstacles in the age of big data.
To ensure efficient machine learning as a service schemes while protecting data
confidentiality, in this work, we propose blind UNET (BUNET), a
secure protocol that implements privacy-preserving medical image
segmentation based on the UNET architecture.  {\song{In BUNET, we
efficiently utilize cryptographic primitives such as homomorphic encryption
and garbled circuits (GC) to design a complete secure protocol for the UNET
neural architecture. In addition, we perform extensive architectural search
in reducing the computational bottleneck of GC-based secure activation protocols
with high-dimensional input data.}}
In the experiment, we thoroughly examine the parameter space of our protocol,
and show that we can achieve up to 14x inference time reduction compared to
the-state-of-the-art secure inference technique on a baseline architecture
with negligible accuracy degradation.

%% file: tex/intro.tex
\section{Introduction}\label{sec:intro}
The use of neural-network (NN) based machine learning (ML) algorithms in
aiding medical diagnosis, especially in the field of medical image computing,
appears to be extremely successful in terms of its prediction 
accuracy. However, the security regulations over medical records contradicts 
the use of big data
in the age of ML.  Highly sensitive patient records are protected
under the Health Insurance Portability and Accountability Act (HIPAA), where
strong protection measures need to be taken over all of the electronic
Protected Health Information (ePHI) possessed by a patient.  In
particular, access controls and client-side encryption are mandated for the
distribution of all ePHI records over public
networks~\cite{hoffman2007securing, HIPAA_Priv}. In addition,
while qualified
professionals are allowed to handle ePHI, the data exposure is required to be
kept minimal~\cite{drolet2017electronic, HIPAA_Mini}, i.e., just enough to
accomplish the necessary professional judgements.

A central question to the real-world deployment of NN-based ML techniques in medical
image processing is how the related data transfer and computations can be
handled securely and efficiently.
Previous security measures on medical data generally involved
physical means (e.g., physically disconnected from the internet), and these
techniques clearly cannot benefit from the large-scale distributed computing networks
available for solving ML tasks.
Recent
advances in cryptography and multi-party secure computing seek alternatives to
address the security concerns. 
In particular, the concept of secure 
inference (SI) is proposed, where
Alice as a client wishes to inference on some of her inputs
with the machine learning models provided by the server, called Bob.  
The security requirement is that no one, including Bob,
learns anything about the inputs from Alice, while Alice also learns nothing
about the models from Bob.  Over the past few years,
prior arts on SI flourished~\cite{liu2017oblivious, mohassel2017secureml,
rouhani2018deepsecure, juvekar2018gazelle, riazi2019xonn, brutzkus2019low},
where secure protocols targeted on general learning problems were proposed. In
addition, we also observe protocol- and system-level
optimizations~\cite{riazi2019xonn, bian2020nass} on SI.  Unfortunately, most
existing works mentioned above do not have a clear application in mind. Thus,
the utilized network architectures and datasets (e.g., MNIST, CIFAR-10) are
usually generic, without immediate practical implications.  

In this work, we propose $\bseg$, a secure protocol for the UNET
architecture~\cite{ronneberger2015u} that enables input-hiding segmentation on
medical images.  In the proposed protocol, we use a combination of cryptographic
building blocks to ensure
that client-side encryption is enforced on all data related to the patients,
and that practical inference time can also be achieved.  
As a result, medical institutions can take advantage of third-party
machine learning service providers without violating privacy regulations.
The main contributions of this work are
summarized as follows.

\begin{itemize}
 \setlength{\itemsep}{0pt}
  \item {\bf{Privacy-Preserving Image Segmentation}}:
    To the best of our knowledge, we are the first to propose a secure
    protocol for image segmentation. 
    
  \item {\bf{Architectural Exploration for Secure UNET}}:
    We perform a search on the possible alternative UNET architectures
    to reduce the amount of computations (in terms of cryptographic 
    realizations) in SI. 

  \item {\bf{Thorough Empirical Evaluations}}:
    By performing architectural-protocol co-design, we achieved 8x--14x
    inference time reduction with negligible accuracy degradation.
\end{itemize}



%% file: tex/background.tex
\section{Preliminaries}\label{sec:background}

\subsection{Cryptographic Primitives}\label{sec:prelim}
In this work, we mainly consider the four types of cryptographic primitives:
a packed additive homomorphic encryption (PAHE) scheme based on the 
ring learning with errors (RLWE) problem~\cite{brakerski2012fully,
fan2012somewhat, brakerski2012leveled, cheon2017homomorphic}, additive
secret sharing (ASS)~\cite{damgaard2016communication}, garbled circuits
(GC)~\cite{yao1982protocols}, and multiplication 
triples (MT)~\cite{beaver1991efficient, keller2018overdrive}. In what
follows, we provide a brief overview for each primitive.

{\bf{PAHE}}: A PAHE is a cryptosystem, where the encryption ($\Enc$) and
decryption ($\Dec$) functions act as group (additive) homomorphisms between
the plaintext and ciphertext spaces. Except for the normal $\Enc$ and
$\Dec$, a PAHE scheme is equipped with the following three
abstract operators. We use $[\bx]$ to denote the encrypted ciphertext of
$\bx\in\zz^{n}$, and $n\in\zz$ the maximum number of plaintext integers that
can be held in a single ciphertext.
\begin{itemize}
  \item Homomorphic addition $(\boxplus)$: for $\bx, \by\in\zz^{n}$, 
    $\Dec([\bx]\boxplus[\by])=\bx+\by$. Note we can also perform homomorphic
    subtraction $\boxminus$, where $\Dec([\bx]\boxminus[\by])=\bx-\by$.
  \item Homomorphic Hadamard product $(\boxcircle)$: for $\bx, \by\in\zz^{n}$,
    $\Dec(\ElementMult{[\bx]}{\by})=\bx\circ \by$, where $\circ$ is the
    element-wise multiplication operator.
  \item Homomorphic rotation $(\rot)$: for $\bx\in\zz^{n}$, let
  $\bx=(x_0, x_1, \cdots, x_{n-1})$, 
    $\rot([\bx], k)=(x_{k}, x_{k+1}, \cdots, x_{n-1}, x_{0}, \cdots, x_{k-1})$
  for $k\in\{0, \cdots, n-1\}$.
\end{itemize}

{\bf{ASS and Homomorphic Secret Sharing}}:
A two-party ASS scheme consists of two operators, $(\share, \rec)$, and some
prime modulus $\pass\in\zz$. Each
operator takes two inputs, where we have
$\bs_{A}=\share(\bx,
\bs_{B})=(\bx-\bs_{B})\bmod{\pass}$ and $\bx=\rec(\bs_{A},
\bs_{B})=(\bs_{A}+\bs_{B})\bmod{\pass}$. In~\cite{juvekar2018gazelle}, homomorphic 
secret sharing (HSS) is adopted, where ASS operates over an encrypted $\bx$.
For HSS, we have that 
\begin{align}
  [\bs_{A}]&=\share([\bx], \bs_{B})=([\bx]\boxminus \bs_{B})\bmod{\pass}
  \text{, and }&\\
  [\bx]&=\rec([\bs_{A}], \bs_{B})=(\bs_{A}\boxplus \bs_{B})\bmod{\pass}.&
\end{align}

{\bf{GC}}: GC can be considered as a more general form of HE. In particular,
the circuit garbler, Alice, ``encrypts'' some function $f$ along with her input
$x$ to Bob, the circuit evaluator. Bob evaluates $f(x, y)$ using his encrypted
input $y$ that is received from Alice obliviously, and obtains the encrypted
outputs. Alice and Bob jointly ``decrypt'' the output of the function $f(x,
y)$ and one of the two parties learns the result.

{\bf{MT}}: Beaver's MT~\cite{beaver1991efficient} is a technique that performs
multiplication on a pair of secret-shared vectors $\bx=\rec(\bs_{A, x},\bs_{B,
x})$ and $\by=\rec(\bs_{A, y}+\bs_{B, y})$ between Alice and Bob. Here, we take
computations performed by Alice as an example, and only note that the exact  same procedure is also executed by Bob on his shares of secrets.
To compute
$\bx\circ \by$, Alice and Bob first pre-share a set of respective
multiplication triples $(\ba_{A}, \bb_{A}, \bc_{A})$ and $(\ba_{B}, \bb_{B},
\bc_{B})$, where we have $\rec(\ba_{A}, \ba_{B})\circ\rec(\bb_{A},
\bb_{B})=\rec(\bc_{A}, \bc_{B})$. 
Alice locally calculates $\bd_{A} = \bs_{A, x}\circ \ba_{A}\bmod{\pass}, \be_{A} =
\bb_{A}\circ \bs_{A, y}\bmod{\pass}$.
Then, Alice and Bob publish their results $\bd_{A}, \bd_{B}$ and $\be_{A},
\be_{B}$.  Finally, Alice obtains 
\begin{align}
  \bg_{A}=\big(\bc_{A}+(\be\circ \bs_{A, 0, i})+(\bd\circ\bs_{A, 1,
  i})-\bd\circ\be\big)\bmod{\pass},
\end{align}
where $\bd=\rec(\bd_{A},\bd_{B})$ and similarly for $\be$.  MT guarantees that
$\rec(\bg_{A}, \bg_{B})=\bx\circ\by$, where $\bg_{B}$ is the MT results
computed by Bob.

\subsection{Related Works on Secure Neural Network Inference}\label{sec:si}
While a limited number of pioneering works have been proposed for secure
inference and training with neural networks~\cite{liu2017oblivious,
mohassel2017secureml, rouhani2018deepsecure, juvekar2018gazelle}, 
it was not until recently that
such protocols carried practical significance. For example,
in~\cite{liu2017oblivious}, an inference with a single CIFAR-10 image takes
more than 500 seconds to complete. Using the same neural architecture, the
performance was improved to less than 8 seconds in one of the most recent
arts on SI, ENSEI~\cite{juvekar2018gazelle}. Unfortunately, as shown
in Section~\ref{sec:experiment}, without UNET-specific optimizations and protocol designs, 
existing approaches carry significant performance overhead, especially on
the 3D images (e.g., CT scans) used in medical applications. 
Hence, in this work, we establish our protocol
based on the ENSEI construction, and explore UNET-specific optimizations and
cryptographic protocol designs to improve the practicality of secure inference 
in medical segmentation.



%% file: tex/propose.tex
\section{Secure UNET for Blind Segmentation}\label{sec:propose}

\subsection{$\bseg$: The Protocol}\label{sec:bseg}
\begin{figure}[!t]
 \centering
 \includegraphics[width=1\columnwidth]{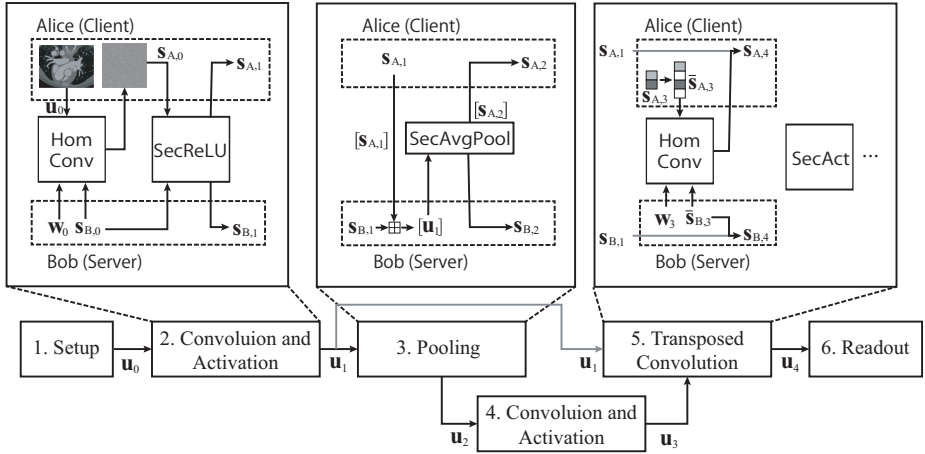}
 \caption{The overview of the $\bseg$ protocol.}
 \label{fig:bunet_gen}
\end{figure}
Fig.~\ref{fig:bunet_gen} shows an example of the $\bseg$ protocol structured
as an UNET architecture, where the input image goes through four
steps. The operators used in each step will be discussed in detail
later in Section~\ref{sec:details}.
\begin{itemize}[leftmargin=*]
  \item[1.] {\bf{Setup}}:
    As our protocol takes both 2D and 3D images as inputs, the input image is
    of dimension $\dim(U)=c\times w\times h\times d$ ($d=1$ for 2D inputs).
    Alice first raster-scans her input image into a
    one-dimensional vector $\bu$ of length $\ell=c\cdot w\cdot h\cdot d$. Bob
    does a similar transformation on his filter weights to obtain the
    one-dimensional vector $\bw$. We use $\bu_{0}$ to denote the input from
    Alice at the $0$-th layer, and $\bw_{0}$ for that of Bob.
  \item[2.] {\bf{Convolution and Activation}}:
    We conduct a standard (input-hiding) convolution on the input
    image with activation functions followed. For each convolution layer with
    activation, we run
    \begin{align}
      \bv_{0}=\HomConv(\bu_{0}, \bw_{0}), \text{ and }
      \bu_{1}=f_{a}(\bv_{0})=\Act(\bv_{0}).
    \end{align}
    for some weight vector $\bw_{0}$. Here, $f_{a}$ is some abstract activation
    function (e.g., ReLU or square activation). The output $\bu_{1}=f_{a}(\bv_{0})$
    will become the input to the next layer. 
    We point out that the activation function
    has a significant impact on both the accuracy and the inference time of UNET.
    Therefore, we propose a hybrid UNET architecture, where both ReLU and square 
    activations are used.
    Here, the security
    guarantee is that Alice obtains no knowledge on $\bw_{0}$, and Bob knows
    nothing about $\bu_{0}$ and $\bv_{0}$ after the protocol execution.
  \item[3.] {\bf{Pooling}}:
    While the standard UNET architecture employs max pooling~\cite{ronneberger2015u} as the
    downsampling method, in the experiment, we show a large performance difference
    between max pooling and average pooling protocols, due to the change of underlying
    protocol. We also demonstrate in Table~\ref{tab:accuracy} that the two pooling 
    methods result in marginal accuracy differences. Hence, we modified the UNET
    architecture to employ only average pooling. Consequently, the proposed
    protocol executes $\bu_{2} = \avgpool(\bu_{1})$. For an input of length
    $\ell$ and pooling size $\zeta_{p}=z_{w, p}\times z_{h, p}\times z_{d,
    p}$, the pooled output have a dimension of $\ell/\zeta_{p}$.
  \item[4.] {\bf{Bottom-Level Convolution and Activation}}: Here, Step 2 is
    repeated, and we get $\bu_{3}$ as output. 
    Note that the input and
    output dimension is reduced by Step 3, so Step 4 is computationally lighter
    than Step 2.
  \item[5.] {\bf{Transposed Convolution and Concatenation}}: 
    While the arithmetic procedures for transposed convolution is essentially
    the same as a normal convolution, protocol-level modifications are
    required for the image concatenation and padding operations. Concretely,
    after obtaining input from the previous layer, e.g., $\bu_{3}$, Alice needs to
    zero-pad $\bu_{3}$ in an interleaving manner, according
    to some stride size $\zeta_{t}=z_{w, t}\times z_{h, t}\times z_{d, t}$.
    The padded result, $\ovbu_{3}$, will have a length of $\ell\cdot
    \zeta_{t}/\zeta_{p}$. Subsequently, Alice
    uses $\ovbu_{3}$ as input to execute the following protocols.
    \begin{align}
      \bv_{4}=\HomConv(\ovbu_{3}, \bw_{3}), \text{ and }
      \bu_{4}=\bv_{4}||\bu_{1}.
    \end{align}
    One subtlety is that, the output from the $1$-st layer, $\bu_{1}$,
    needs to be concatenated with the output from the transposed convolution
    layer for the rest of the normal convolutions.  However, $\bu_{1}$
    and $\bv_{4}$ will actually be encrypted under different keys.
    Thus, both results need to be decrypted and concatenated by Alice.
    The concatenated
    result, $\bv_{1} || \bu_{4}$, will become the inputs to later layers.
  \item[6.] {\bf{Readout}}: 
    Here, we applies the $\secargmax$ function over the label dimensions. It
    is noted that, since the ${\mathsf{Softmax}}$ operator is monotonic and is
    only required in the learning process, we avoid using a separate protocol
    for ${\mathsf{Softmax}}$, and directly perform a secure
    ${\mathsf{Argmax}}$. Since the $\mathsf{Argmax}$ function is a pixel-wise
    comparison function across the label dimension, it can be implemented
    using a simple GC protocol similar to the secure ReLU protocol, and 
    we omit a formal presentation.
\end{itemize}

\subsubsection{Threat Model and Security}
The threat model for $\bseg$ is that both Bob
and Alice are semi-honest, in the sense that both parties follow the 
protocol prescribed above (e.g., encrypting real data with $\Enc$, etc.), 
but want to
learn as much information as possible from the other party. Our protocol 
guarantees that Alice learns only the segmentation results, while Bob learns
nothing about the inputs from Alice. 

The security of the proposed protocol can be easily reduced to that of existing 
works~\cite{juvekar2018gazelle, bian2020ensei}, where any attack against 
$\bseg$ will result in a non-negligible advantage against the ENSEI~\cite{bian2020ensei} 
and Gazelle~\cite{juvekar2018gazelle} protocols.

\subsection{The Cryptographic Building Blocks}\label{sec:details}
Here, we discuss each of the cryptographic primitives used in the previous
section in details.

{\bf{$\HomConv$}}: The $\HomConv$ operator obliviously convolve two
vectors $\bu$ and $\bw$. 
A very recent
work~\cite{bian2020ensei} discovered that, instead of the complex
rotate-and-accumulate approach proposed by previous
works~\cite{juvekar2018gazelle}, homomorphic convolution can be performed in
the frequency domain, where the only computation needed in the homomorphic
domain is the $\boxcircle$ (homomorphic Hadamard product) operator. Therefore,
the homomorphic convolution protocol proceeds as follows.
\begin{itemize}[leftmargin=*]
  \item[1.] First, Alice performs an integer discrete Fourier transform (DFT)
    (i.e., number theoretic transform in~\cite{bian2020ensei}) on $\bu$ and
    obtain its frequency-domain representation, $\hatbu$. She simply encrypts
    this input array into a ciphertext $[\hatbu]$ (when $\ell>n$, the vector
    is encrypted into multiple ciphertexts)
    by running $[\hatbu]=\Enc_{\crk}(\hatbu)$, where $\crk$ is
    the encryption key. The resulting ciphertext is transferred to Bob.
  \item[2.] Before receiving any input from Alice, Bob applies DFT on his filter
    weights to obtain $\hatbw$. In this process, the size of the filter will be padded to be
    $\ell$. Upon receiving the inputs $[\hatbu]$ from Alice, 
    Bob computes
    \begin{align}
      [\hatbv] = [\hatbu\circ\hatbw] = [\hatbu]\boxcircle \hatbw
    \end{align}
    for all ciphertexts $\hatbu$.
  \item[3.] Finally, since $\hatbv$ contains information of the weights from Bob,
    Bob applies HSS as $[\hatbs_{A}]=\share([\hatbv], \hatbs_{B})$.  Bob keeps 
    $\hatbs_{B}$ and returns
    $[\hatbs_{A}]$ to Alice, where Alice decrypts and obtain $\hatbs_{A}$. 
    Both Alice and Bob run inverse DFT on their shares of secrets (i.e.,
    $\hatbs_{A}$ and $\hatbs_{B}$) and obtain $\bs_{A}$ and $\bs_{B}$, 
    respectively, completing the protocol.
  \end{itemize}

{\bf{$\Act$}}: The $\Act$ protocols are summarized as follows.
\begin{itemize}[leftmargin=*]
  \item {\bf{ReLU}}: We follow the construction in~\cite{juvekar2018gazelle}
    based on the GC protocol. Alice first garbles the circuit 
    with her share of secret $\bs_{A}$. The garbled circuit
    obliviously computes the following function
    \begin{align}
      \bv=\rec(\bs_{A}, \bs_{B}), 
      {\text{ and }} \ovbs_{A} = \share(\mathsf{ReLU}(\bv),  \ovbs_{B})\label{eq:gc_relu}
    \end{align}
    where $\ovbs_{B}$ is a freshly generated share of secret from Bob.
    After protocol execution, Alice obtains $\ovbs_{A}$, which 
    contains $\mathsf{ReLU}(\bv)$ in an oblivious manner.

  \item {\bf{Square}}: Since the square activation (i.e., $y=x^{2}$) is
    essentially evaluating a polynomial over the inputs, the
    computationally-light MT can be used instead of GC. To use MT, we first
    share the secret $\bv$ twice among Alice and Bob.
    Then, the MT protocol outlined in Section~\ref{sec:prelim} can be
    executed, where both $\bx$ and $\by$ equal $\bv$. After the protocol
    execution, Alice and Bob respectively obtain $\bg_{A}$ and $\bg_{B}$ where
    $\rec(\bg_{A}, \bg_{B})=\bv^{2}$.
    The main observation here is that, all computations are coefficient-wise
    multiplications and additions over $\lg{\pass}$-bit integers. As a result, 
    MT-based square activation is much faster than GC-based ReLU activation.

\end{itemize}

{\bf{$\mathsf{SecPool}$}}: As mentioned above, two types of secure pooling can
be implemented, the $\mathsf{SecMaxPool}$ and the $\avgpool$ operator.
In~\cite{juvekar2018gazelle}, it is shown that max-pooling can be implemented
using the GC protocol as in Eq.~\eqref{eq:gc_relu}, where we replace the
$\mathsf{ReLU}$ operator with the $\mathsf{MaxPool}$ operator. Meanwhile, for
secure average pooling, we can use a simple protocol that is
purely based on PAHE. Specifically, we can compute the window-wise
sum of some vector $\bv$ by calculating
$\avgpool(\bv)=\sum_{i=0}^{\zeta_{p}-1}\rot([\bv], i)$, where $\zeta_{p}$ is
the pooling window size. Since both homomorphic rotations and additions are
light operations compared to GC, $\avgpool$ is much faster than
$\mathsf{SecMaxPool}$.

%% file: tex/result.tex
\section{Accuracy Experiments and Protocol Instantiation}\label{sec:experiment}

\subsection{Experiment Setup}
Due to the lack of immediate existing works, we compare $\bseg$ with
the standard UNET architecture implemented by the
ENSEI~\cite{bian2020ensei} protocol, which is the best 
performing protocol on secure
multi-class inference. Here, the standard UNET architecture 
only utilizes max-pooling for pooling layers, and
ReLU for activation layers. We denote this architecture as
the baseline architecture.  As shown in  the appendix, the baseline architecture consists 
19 convolution layers including three transposed
convolution layers, 14 activation layers, three average pooling 
layers, and a readout layer implementing the $\secargmax$ function.

Our experiments 
are conducted on three datasets, GM~\cite{prados2017spinal},
EM~\cite{cardona2010integrated}, and HVSMR~\cite{pace2015interactive}.  Due to
the space constraint, we only present the accuracy and performance results on
the EM (two dimensional) and HVSMR (three dimensional) 
datasets (GM will be added to the appendix).


The cryptographic performance is characterized on an Intel i5-9400 2.9\,GHz
CPU, and the accuracy results are obtained with an NVIDIA P100 GPU.  The
adopted PAHE library is SEAL version 3.3.0~\cite{sealcrypto} (we set
$q, p$ to be 60-bit and 20-bit integers, respectively, and $n=2048$,
ensuring a 128-bit security) and 
MT/GC protocols are implemented using LowGear provided by
MP-SPDZ~\cite{keller2018overdrive,lowgear}.


\subsection{Accuracy and Performance Results}

\begin{table}[t]
  \scriptsize
  \centering
  \renewcommand\arraystretch{1}
  \caption{The Dice Accuracy on the HVSMR Dataset for Different Neural Architectures}
  \label{tab:accuracy}
  \begin{tabular}{c|c|c|c|c|c|c}
  \hline
                        & ReLU+Max       & ReLU           & ReLU+Max          & ReLU            & Hybrid          & Square             \\
                        & Float          & Float          & 32-bit (Baseline) & 16-bit          & 20-bit          & 16-bit             \\\hline
    HVSMR Myo. Dice          & 0.74$\pm 0.04$ & 0.73$\pm 0.04$ & 0.73$\pm0.05$  & 0.73$\pm0.05$   & 0.71$\pm 0.05$  & 0.52$\pm 0.14$     \\\hline
    HVSMR BP Dice            & 0.87$\pm 0.04$ & 0.87$\pm 0.04$ & 0.87$\pm0.04$  & 0.87$\pm0.04$   & 0.86$\pm 0.05$  & 0.83$\pm 0.05$     \\\hline
    HVSMR Time~(s)  & -                & -              & 42616      & 14205           & 3054.2          & 1118.2             \\\hline\hline
    EM Dice                  & 0.9405  & -              & 0.9411     & 0.9398           & 0.9385         & 0.8767\\\hline
    EM Time~(s)              & -       & -              & 8838.1     & 2968.6           & 1077.9          & 227.16             \\\hline\hline
  \end{tabular}
\end{table}

%
\begin{figure}[t!]
    \centering
     \includegraphics[width=0.6\columnwidth]{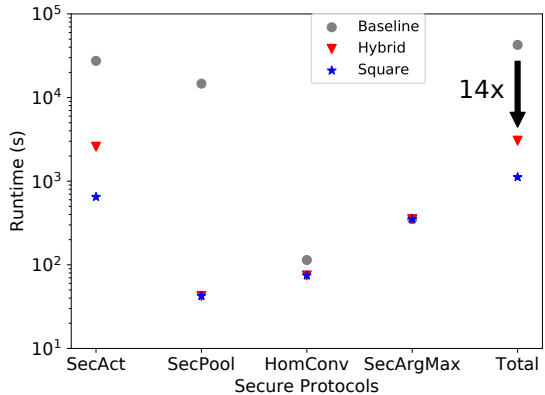}
    \caption{Runtime distribution for different cryptographic building blocks in a single run of secure segmentation.}
    \label{fig:runtime}
\end{figure}

  \begin{table}[!t]
  \scriptsize
  \tabcolsep 10pt
  \centering
  \caption{Number of Activation Functions per Layer Batch}
  \label{tab:perlayer}
  \begin{tabular}{c|c|c|c|c|c|c}
  \hline
    1st & 2nd & 3rd & 4th & 5th & 6th & 7th \\\hline
    33554432 & 8388608 & 2097152 & 524288 & 2097152 & 8388608 & 33554432\\\hline
  \end{tabular}
\end{table}

In this section, we explore how the neural architecture of 3D-UNET impact on
the segmentation performance. 
In particular, it is important to see if
the the proposed architectural modifications for UNET result in satisfactory
prediction accuracy while accelerating the network inference time.
We downsampled the images in the HVSMR~\cite{pace2015interactive} 
dataset to a
dimension of $64\times 64\times 64$ containing three 
class labels, i.e., background, myocardium (myo.) and blood pool (BP). 
The images in the EM dataset has $200\times 200$ input for binary 
segmentation.
Table~\ref{tab:accuracy} summarizes the dice scores 
and runtime under various architectural
settings (HVSMR with 10-fold cross validation). 
Here, the pooling function is average pooling unless otherwise
stated. Hybrid refers to the neural architecture where the first and the last
layer batches use square activation, 
while all other layer batches adopt ReLU activation. Here, a layer batch denotes 
two convolution and activation layers with the same output feature dimensions and 
channels. 
We have three main
observations. First, the use of average pooling instead of max pooling results
in negligible accuracy degradation, on a level that can likely be compensated by
parameter tuning. Second, the UNET architecture is robust in low-quantization
environment, where we see little accuracy difference between floating
point, 32-bit and 16-bit quantization factors, especially on the EM dataset. 
Lastly, replacing all ReLU
activations with squares results in significant accuracy degradation for
the segmentation of myocardium. However, the BP
prediction can be acceptable for a quick run of cheap evaluations,
and the hybrid architecture successfully achieves
a good balance between accuracy and performance.

We record the total runtime for a single blind image segmentation with respect to
different 3-D UNET architectures. In Table~\ref{tab:perlayer} and Fig.~\ref{fig:runtime}, 
we illustrate the per-layer-batch number of neuron activations and the runtime 
distribution for different secure protocols in $\bseg$. 
As expected, the first and last two layers contain the most amount of
activations, and replacing the GC-based heavy ReLU activation with square activation 
results in an immediate 5x total runtime reduction. 
In addition, it is observed that
the runtime for ReLU activation functions dominate the total runtime across architectures,
while square activation is as light as a frequency-domain homomorphic convolution 
operation. 

Compared to the baseline 32-bit ReLU architecture, we obtain 8x--14x runtime 
reduction with the reasonably accurate Hybrid architecture,
and up to 39x reduction with the cheapest (all-square) UNET implementation on EM.
Finally, we note that most NN operations can be parallelized as
well as the cryptographic building blocks. Therefore, since our
performance is recorded on a single-thread CPU, we expect
further runtime reduction for $\bseg$ on parallel
computing architectures.

%% file: tex/conclusion.tex
\section{Conclusion}\label{sec:conclusion}
In this work, we propose {\bseg} to perform blind medical image segmentation
on encrypted medical images. The observation we make is that protocol and network
designs need to be jointly performed to achieve the best accuracy-performance
trade-off. By designing UNET-specific protocols
and optimizing the UNET architecture, we show that up to 8x--14x inference time 
reduction can be achieved with negligible accuracy degradation on several 
medical datasets.

%% file: tex/ack.tex
\section*{Acknowledgment}
\sloppy
This work was partially supported by JSPS KAKENHI Grant No.~20K19799, 20H04156, Edgecortix Inc, the Science and Technology Planning Project of Guangdong Province under Grant No. 2017A070701013, 2017B090904034, 2017B030314109, 2018B090944002, and 2019B020230003, Guangdong peak project under Grant No. DFJH201802, the National Key Research and Development Program under Grant No. 2018YFC1002600, the Natural Science Foundation of Guangdong Province under Grant No. 2018A030313785.

%% file: appendix/appendix.tex
\title{
  Appendix for BUNET: Blind Medical Image Segmentation Based on Secure UNET
}

\author{
Song Bian \inst{1} \and 
Xiaowei Xu \inst{2}  \and 
Weiwen Jiang \inst{3} \and
Yiyu Shi\inst{3} 
Takashi Sato\inst{1}
}
\authorrunning{S. Bian, et al.}
\institute{
Kyoto University\\
\email{\{sbian, takashi\}@easter.kuee.kyoto-u.ac.jp}
\and Guangdong Provincial People's Hospital\\
\email{xiao.wei.xu@foxmail.com}
\and University of Notre Dame\\
\email{{\{wjiang2, yshi4\}@nd.edu}}
}

\maketitle

\setlength{\textfloatsep}{6pt}
\setlength{\floatsep}{6pt}
\setlength{\dbltextfloatsep}{6pt}

\section{Additional Experiment Results for EM/GM datasets}
\begin{table}[h]
  \scriptsize
  \centering
  \renewcommand\arraystretch{1}
  \caption{The Dice Accuracy and Runtime for the GM Dataset for Different Neural Architectures. Note that since we have the same architecture and input dimensions
  for GM and EM, the runtime for the secure inference protocol is the same.}
  \label{tab:accuracy}
  \begin{tabular}{c|c|c|c|c|c|c}
  \hline
                        & ReLU+Max       & ReLU           & ReLU+Max          & ReLU            & Hybrid          & Square             \\
                        & Float          & Float          & 32-bit (Baseline) & 16-bit          & 16-bit          & 16-bit             \\\hline
    GM Dice                  & 0.5632  & -              & 0.5581     & 0.5577           & 0.5607         & 0.5546\\\hline
    GM Time~(s)              & -       & -              & 8838.1     & 2968.6           & 1077.9          & 227.16             \\\hline
  \end{tabular}
\end{table}
\begin{figure}[h]
 \centering
 \includegraphics[width=0.5\columnwidth]{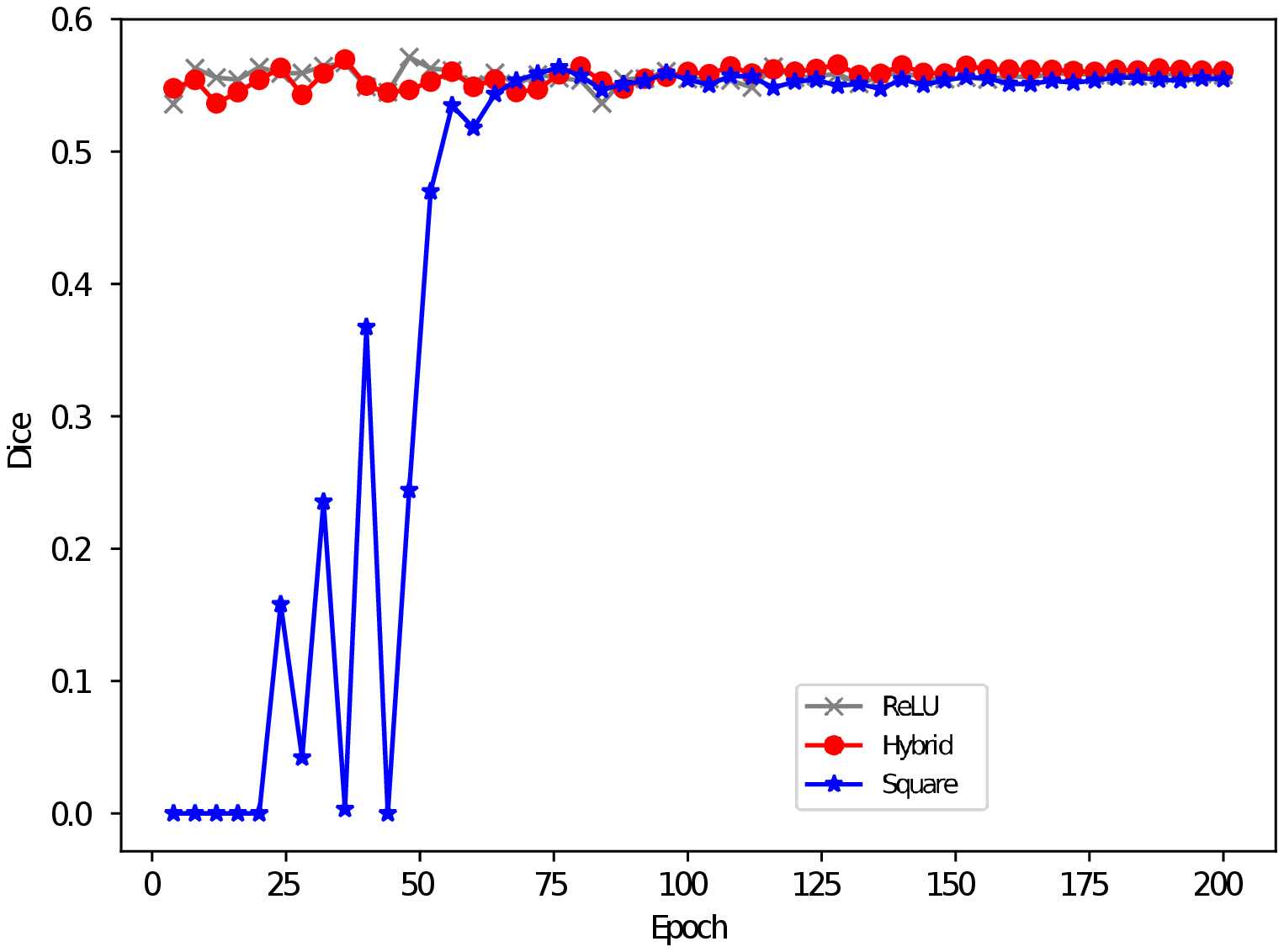}
 \caption{The learning curve for different neural architectures over 200 epoches. It
 can be observed that the neural architecture with all-square activation encounters
 some difficulties initially, but successfully proceeded learning later on.}
 \label{fig:bunet_gen}
\end{figure}

\section{Detailed Network Architecture}
\begin{table}[h]
  \scriptsize
  \centering
  \renewcommand\arraystretch{1}
  \caption{The neural architecture for the 3-D UNET used in HVSMR dataset. 
  The input is of dimension $1\times 64\times 64\times 64=64^{3}$ where $1$ denotes a single channel. For EM and GM, we replace the input dimension $64^{3}$ with $200^{2}$,
  and the filter dimensions are kept the same. Here, Quant. 
  and Quantization indicate
  a quantization layer that quantize all the outputs from the
  previous operation (e.g., convolution or square activation).
  Meanwhile, Norm. denotes a normalization layer. }
  \label{tab:accuracy}
  \begin{tabular}{c|c|c|c|c}
  \hline
  Layer Batch & Layer Type & Input Dim.       & Filter Dim. & Output Dim. \\\hline
  1           & Convolution+Quant.    & $1\times 64^{3}$ & $64\times 3^{3}$    & $64\times 64^{3}$ \\\hline
  1           & Activation+Norm.     & $64\times 64^{3}$         & -           & $64\times 64^{3}$ \\\hline
  1           & Quantization                 &  $64\times 64^{3}$         & -           & $64\times 64^{3}$ \\\hline
  1           & Convolution+Quant.    & $64\times 64^{3}$ & $64\times 3^{3}$   & $64\times 64^{3}$ \\\hline
  1           & Activation+Norm.     & $64\times 64^{3}$         & -           & $64\times 64^{3}$ \\\hline
  1           & Quantization                 &  $64\times 64^{3}$         & -           & $64\times 64^{3}$ \\\hline
  1           & Pooling        & $64\times 64^{3}$         & $2^{3}$      & $64\times 32^{3}$ \\\hline
  2           & Convolution+Quant.    & $64\times 32^{3}$ & $128\times 3^{3}$    & $128\times 32^{3}$ \\\hline
  2           & Activation+Norm.     & $128\times 32^{3}$         & -           & $128\times 32^{3}$ \\\hline
  2           & Quantization                 &  $128\times 32^{3}$         & -           & $128\times 32^{3}$ \\\hline
  2           & Convolution+Quant.    & $128\times 32^{3}$ & $128\times 3^{3}$   & $128\times 32^{3}$ \\\hline
  2           & Activation+Norm.     & $128\times 32^{3}$         & -           & $128\times 32^{3}$ \\\hline
  2           & Quantization                 &  $128\times 32^{3}$         & -           & $128\times 32^{3}$ \\\hline
  2           & Pooling        & $128\times 32^{3}$         & $2^{3}$      & $128\times 16^{3}$ \\\hline
  3           & Convolution+Quant.    & $128\times 16^{3}$ & $256\times 3^{3}$    & $256\times 16^{3}$ \\\hline
  3           & Activation+Norm.     & $256\times 16^{3}$         & -           & $256\times 16^{3}$ \\\hline
  3           & Quantization                 &  $256\times 16^{3}$         & -           & $256\times 16^{3}$ \\\hline
  3           & Convolution+Quant.    & $256\times 16^{3}$ & $256\times 3^{3}$   & $256\times 16^{3}$ \\\hline
  3           & Activation+Norm.     & $256\times 16^{3}$         & -           & $256\times 16^{3}$ \\\hline
  3           & Quantization                 &  $256\times 16^{3}$         & -           & $256\times 16^{3}$ \\\hline
  3           & Pooling        & $256\times 16^{3}$         & $2^{3}$      & $256\times 8^{3}$ \\\hline 
  4           & Convolution+Quant.    & $256\times 8^{3}$ & $512\times 3^{3}$    & $512\times 8^{3}$ \\\hline
  4           & Activation+Norm.     & $512\times 8^{3}$         & -           & $512\times 8^{3}$ \\\hline
  4           & Quantization                 &  $512\times 8^{3}$         & -           & $512\times 8^{3}$ \\\hline
  4           & Convolution+Quant.    & $512\times 8^{3}$ & $512\times 3^{3}$   & $512\times 8^{3}$ \\\hline
  4           & Activation+Norm.     & $512\times 8^{3}$         & -           & $512\times 8^{3}$ \\\hline
  4           & Quantization                 &  $512\times 8^{3}$         & -           & $512\times 8^{3}$ \\\hline
  4           & Transposed Convolution+Quant.    & $512\times 8^{3}$ & $256\times 2^{3}$   & $256\times 16^{3}$ \\\hline
  5           & Convolution+Quant.    & $512\times 16^{3}$ & $256\times 3^{3}$    & $256\times 16^{3}$ \\\hline
  5           & Activation+Norm.     & $256\times 16^{3}$         & -           & $256\times 16^{3}$ \\\hline
  5           & Quantization                 &  $256\times 16^{3}$         & -           & $256\times 16^{3}$ \\\hline
  5           & Convolution+Quant.    & $256\times 16^{3}$ & $256\times 3^{3}$   & $256\times 16^{3}$ \\\hline
  5           & Activation+Norm.     & $256\times 16^{3}$         & -           & $256\times 16^{3}$ \\\hline
  5           & Quantization                 &  $256\times 16^{3}$         & -           & $256\times 16^{3}$ \\\hline
  5           & Transposed Convolution+Quant.    & $256\times 16^{3}$ & $128\times 2^{3}$   & $128\times 32^{3}$ \\\hline
  6           & Convolution+Quant.    & $256\times 32^{3}$ & $128\times 3^{3}$    & $128\times 32^{3}$ \\\hline
  6           & Activation+Norm.     & $128\times 32^{3}$         & -           & $128\times 32^{3}$ \\\hline
  6           & Quantization                 &  $128\times 32^{3}$         & -           & $128\times 32^{3}$ \\\hline
  6           & Convolution+Quant.    & $128\times 32^{3}$ & $128\times 3^{3}$   & $128\times 32^{3}$ \\\hline
  6           & Activation+Norm.     & $128\times 32^{3}$         & -           & $128\times 32^{3}$ \\\hline
  6           & Quantization                 &  $128\times 32^{3}$         & -           & $128\times 32^{3}$ \\\hline
  6           & Transposed Convolution+Quant.    & $128\times 32^{3}$ & $64\times 2^{3}$   & $64\times 64^{3}$ \\\hline
  7           & Convolution+Quant.    & $128\times 64^{3}$ & $64\times 3^{3}$    & $64\times 64^{3}$ \\\hline
  7           & Activation+Norm.     & $64\times 64^{3}$         & -           & $64\times 64^{3}$ \\\hline
  7           & Quantization                 &  $64\times 64^{3}$         & -           & $64\times 64^{3}$ \\\hline
  7           & Convolution+Quant.    & $64\times 64^{3}$ & $64\times 3^{3}$   & $64\times 64^{3}$ \\\hline
  7           & Activation+Norm.     & $64\times 64^{3}$         & -           & $64\times 64^{3}$ \\\hline
  7           & Quantization                 &  $64\times 64^{3}$         & -           & $64\times 64^{3}$ \\\hline
  8           & Convolution    & $64\times 64^{3}$ & $3\times 3^{3}$   & $3\times 64^{3}$ \\\hline
  9           & Argmax         & $3\times 64^{3}$ &  -                 & $1\times 64^{3}$ \\\hline
  \end{tabular}
\end{table}


%% file: main.bbl
\begin{thebibliography}{10}
\providecommand{\url}[1]{\texttt{#1}}
\providecommand{\urlprefix}{URL }
\providecommand{\doi}[1]{https://doi.org/#1}

\bibitem{beaver1991efficient}
Beaver, D.: Efficient multiparty protocols using circuit randomization. In:
  Annual International Cryptology Conference. pp. 420--432. Springer (1991)

\bibitem{bian2020nass}
Bian, S., Jiang, W., Lu, Q., Shi, Y., Sato, T.: {NASS}: Optimizing secure
  inference via neural architecture search. arXiv preprint arXiv:2001.11854
  (2020)

\bibitem{bian2020ensei}
Bian, S., Wang, T., Hiromoto, M., Shi, Y., Sato, T.: {ENSEI}: Efficient secure
  inference via frequency-domain homomorphic convolution for privacy-preserving
  visual recognition (2020)

\bibitem{brakerski2012fully}
Brakerski, Z.: Fully homomorphic encryption without modulus switching from
  classical {G}ap{SVP}. In: Advances in Cryptology--CRYPTO 2012, pp. 868--886.
  Springer (2012)

\bibitem{brakerski2012leveled}
Brakerski, Z., Gentry, C., Vaikuntanathan, V.: ({L}eveled) fully homomorphic
  encryption without bootstrapping. ACM Transactions on Computation Theory
  (TOCT)  \textbf{6}(3), ~13 (2014)

\bibitem{brutzkus2019low}
Brutzkus, A., Gilad-Bachrach, R., Elisha, O.: Low latency privacy preserving
  inference. In: International Conference on Machine Learning. pp. 812--821
  (2019)

\bibitem{cardona2010integrated}
Cardona, A., Saalfeld, S., Preibisch, S., Schmid, B., Cheng, A., Pulokas, J.,
  Tomancak, P., Hartenstein, V.: An integrated micro-and macroarchitectural
  analysis of the drosophila brain by computer-assisted serial section electron
  microscopy. PLoS biology  \textbf{8}(10) (2010)

\bibitem{cheon2017homomorphic}
Cheon, J.H., Kim, A., Kim, M., Song, Y.: Homomorphic encryption for arithmetic
  of approximate numbers. In: International Conference on the Theory and
  Application of Cryptology and Information Security. pp. 409--437. Springer
  (2017)

\bibitem{damgaard2016communication}
Damg{\aa}rd, I., Nielsen, J.B., Polychroniadou, A., Raskin, M.: On the
  communication required for unconditionally secure multiplication. In: Annual
  International Cryptology Conference. pp. 459--488. Springer (2016)

\bibitem{drolet2017electronic}
Drolet, B.C., Marwaha, J.S., Hyatt, B., Blazar, P.E., Lifchez, S.D.: Electronic
  communication of protected health information: privacy, security, and hipaa
  compliance. The Journal of hand surgery  \textbf{42}(6),  411--416 (2017)

\bibitem{fan2012somewhat}
Fan, J., Vercauteren, F.: Somewhat practical fully homomorphic encryption. IACR
  Cryptology ePrint Archive  \textbf{2012}, ~144 (2012)

\bibitem{HIPAA_Priv}
HHS.gov:
  \url{https://www.hhs.gov/hipaa/for-professionals/privacy/laws-regulations/index.html}
  (2009), accessed: 2020-03-04

\bibitem{HIPAA_Mini}
HHS.gov:
  \url{https://www.hhs.gov/hipaa/for-professionals/privacy/guidance/minimum-necessary-requirement/index.html}
  (2009), accessed: 2020-03-04

\bibitem{hoffman2007securing}
Hoffman, S., Podgurski, A.: Securing the hipaa security rule. Journal of
  Internet Law, Spring pp. 06--26 (2007)

\bibitem{juvekar2018gazelle}
Juvekar, C., et~al.: Gazelle: A low latency framework for secure neural network
  inference. arXiv preprint arXiv:1801.05507  (2018)

\bibitem{keller2018overdrive}
Keller, M., Pastro, V., Rotaru, D.: Overdrive: making spdz great again. In:
  Annual International Conference on the Theory and Applications of
  Cryptographic Techniques. pp. 158--189. Springer (2018)

\bibitem{liu2017oblivious}
Liu, J., et~al.: Oblivious neural network predictions via {M}inio{NN}
  transformations. In: Proc. of ACM SIGSAC Conference on Computer and
  Communications Security. pp. 619--631. ACM (2017)

\bibitem{mohassel2017secureml}
Mohassel, P., et~al.: Secureml: A system for scalable privacy-preserving
  machine learning. In: Proc. of Security and Privacy (SP). pp. 19--38. IEEE
  (2017)

\bibitem{lowgear}
MP-SPDZ: \url{https://github.com/data61/MP-SPDZ/} (2018), accessed: 2020-03-10

\bibitem{pace2015interactive}
Pace, D.F., Dalca, A.V., Geva, T., Powell, A.J., Moghari, M.H., Golland, P.:
  Interactive whole-heart segmentation in congenital heart disease. In:
  International Conference on Medical Image Computing and Computer-Assisted
  Intervention. pp. 80--88. Springer (2015)

\bibitem{prados2017spinal}
Prados, F., Ashburner, J., Blaiotta, C., Brosch, T., Carballido-Gamio, J.,
  Cardoso, M.J., Conrad, B.N., Datta, E., D{\'a}vid, G., De~Leener, B., et~al.:
  Spinal cord grey matter segmentation challenge. Neuroimage  \textbf{152},
  312--329 (2017)

\bibitem{riazi2019xonn}
Riazi, M.S., Samragh, M., Chen, H., Laine, K., Lauter, K.E., Koushanfar, F.:
  Xonn: Xnor-based oblivious deep neural network inference. IACR Cryptology
  ePrint Archive  \textbf{2019}, ~171 (2019)

\bibitem{ronneberger2015u}
Ronneberger, O., Fischer, P., Brox, T.: U-net: Convolutional networks for
  biomedical image segmentation. In: International Conference on Medical image
  computing and computer-assisted intervention. pp. 234--241. Springer (2015)

\bibitem{rouhani2018deepsecure}
Rouhani, B.D., et~al.: Deepsecure: Scalable provably-secure deep learning. In:
  Proc. of DAC. pp.~1--6. IEEE (2018)

\bibitem{sealcrypto}
{M}icrosoft {SEAL} (release 3.3). \url{https://github.com/Microsoft/SEAL} (June
  2019), microsoft Research, Redmond, WA.

\bibitem{yao1982protocols}
Yao, A.C.: Protocols for secure computations. In: Foundations of Computer
  Science, 1982. SFCS'08. 23rd Annual Symposium on. pp. 160--164. IEEE (1982)

\end{thebibliography}
